\documentclass[journal]{IEEEtran}

\IEEEoverridecommandlockouts
\usepackage[dvips]{graphicx}
\usepackage{algorithm}
\usepackage{algorithmic}
\usepackage{psfrag}

\usepackage{bm}
\usepackage{amssymb,mathtools}
\usepackage{amsfonts}
\usepackage{amsmath,amssymb}
\usepackage{amsthm}
\usepackage{cite} 
\usepackage[ancient]{flushend}
\usepackage[dvipsnames]{xcolor}
\usepackage{url}
\usepackage[hyperfootnotes=false,hidelinks]{hyperref}
\usepackage{graphicx}
\usepackage{upgreek}
\usepackage{yfonts}
\usepackage{cleveref}
\usepackage{caption}
\usepackage{amsmath}
\usepackage{cite}
\usepackage{mathtools}
\usepackage{subfigure}
\usepackage{psfrag}
\usepackage{scalerel}
\DeclarePairedDelimiter\ceil{\lceil}{\rceil}

\newtheorem{theorem}{\bf Theorem}

\begin{document}

\title{Asynchronous Online Federated Learning with Reduced Communication Requirements}

\author{Francois Gauthier, \textit{Member}, \textit{IEEE}, Vinay Chakravarthi Gogineni, \textit{Member}, \textit{IEEE}, Stefan Werner, \textit{Senior Member}, \textit{IEEE}, Yih-Fang Huang, \textit{Life Fellow}, \textit{IEEE}, Anthony Kuh, \textit{Fellow}, \textit{IEEE} \vspace{-2ex}
\thanks{This work was supported by the Research Council of Norway.}
\thanks{A conference precursor of this work appears in the 2022 IEEE International Conference on Communications \cite{ICC}.}
\thanks{F. Gauthier, V. C. Gogineni, and S. Werner are with the department of Electronic Systems, Norwegian University of Science and Technology, Trondheim, Norway Email: \{francois.gauthier, vinay.gogineni, stefan.werner\}@ntnu.no.}
\thanks{Yih-Fang Huang is with the Department of Electrical Engineering, University of Notre Dame, Notre Dame, IN 46556 USA (e-mail: huang@nd.edu).}
\thanks{Anthony Kuh is with the Department of Electrical Engineering, University of Hawaii at Manoa, Honolulu, HI 96882 USA (e-mail: kuh@hawaii.edu). Anthony Kuh acknowledges support in part by NSF Grant 2142987}
}

\makeatletter
\newcommand{\linebreakand}{%
  \end{@IEEEauthorhalign}
  \hfill\mbox{}\par
  \mbox{}\hfill\begin{@IEEEauthorhalign}
}
\makeatother

\newcommand{\Lim}[1]{\raisebox{0.5ex}{\scalebox{0.8}{$\displaystyle \lim_{#1}\;$}}}

\maketitle

\begin{abstract}

Online federated learning (FL) enables geographically distributed devices to learn a global shared model from locally available streaming data. Most online FL literature considers a best-case scenario regarding the participating clients and the communication channels. However, these assumptions are often not met in real-world applications. Asynchronous settings can reflect a more realistic environment, such as heterogeneous client participation due to available computational power and battery constraints, as well as delays caused by communication channels or straggler devices. Further, in most applications, energy efficiency must be taken into consideration. Using the principles of partial-sharing-based communications, we propose a communication-efficient asynchronous online federated learning (PAO-Fed) strategy. By reducing the communication overhead of the participants, the proposed method renders participation in the learning task more accessible and efficient. In addition, the proposed aggregation mechanism accounts for random participation, handles delayed updates and mitigates their effect on accuracy. We prove the first and second-order convergence of the proposed PAO-Fed method and obtain an expression for its steady-state mean square deviation. Finally, we conduct comprehensive simulations to study the performance of the proposed method on both synthetic and real-life datasets. The simulations reveal that in asynchronous settings, the proposed PAO-Fed is able to achieve the same convergence properties as that of the online federated stochastic gradient while reducing the communication overhead by 98 percent.

\end{abstract}


\section{Introduction}
A myriad of intelligent devices, such as smartphones, smartwatches, and connected refrigerators, are becoming an integral part of our daily lives, and an enormous amount of data is available on those devices. Unfortunately, this data is primarily unused, and we need to develop the tools that can process this data to extract information that can improve our daily lives while, at the same time, ensuring our privacy. Federated learning (FL) \cite{FedOverview} provides an adaptive large-scale collaborative learning framework suitable for this task. In FL, a server aggregates information received from distributed devices referred to as clients to train a global shared model; the clients do not share any private data with the server, only their local model parameters or gradients learned from this data \cite{origin, FedOverview}. When data becomes continuously available to clients, and we want to learn from it in real-time, we can utilize online FL \cite{OnlineOrigin}, e.g., sensor monitoring \cite{FLapplications}. In online FL, the server aggregates the local models learned on the streaming data of the clients \cite{FLstreaming}. However, in many applications, the participating clients might have heterogeneous energy and communication capacity and can be intermittently unavailable or subject to failure. Therefore, such edge devices cannot participate in typical federated learning implementations.

In most real-world implementations of FL, it is essential to consider statistical heterogeneity, system heterogeneity, and imperfect communication channels between clients and the server. Statistical heterogeneity implies that data are imbalanced and non-i.i.d. \cite{Noniid} across devices, while system heterogeneity refers to their various computational and communication capacities. Finally, imperfect communication channels cause delays in the exchanged messages. Although many FL approaches can handle statistical heterogeneity, there is relatively little research addressing the remaining complications above. In particular, existing FL methods commonly assume a best-case scenario concerning the client availability and performance as well as perfect channel conditions \cite{CommEnergyFL, CommBasicFL, Accelerated, FedOverview, FedAvgCV, Compression, Energy, FLCompress1,FLCompress2,FLCompress3, FedAVG}. However, several additional aspects need attention for efficient FL in a realistic setting. First, clients cannot be expected to have the same participation frequency, e.g.,  due to diverse resource constraints, channel availability, or concurrent solicitations \cite{Async,Async2, AsyncOnline,AsyncCommSched}. Furthermore, clients may become unavailable for a certain period during the learning process, i.e., some clients are malfunctioning or not reachable by the server \cite{Async, Async2}. In addition, physical constraints such as distance or overload introduce delays in the communication between the clients and the server, making their contribution arrive later than expected \cite{Async2, AsyncOnline, AsyncTier,AsyncCommSched}. These constraints, frequently occurring in practice, impair the efficiency of FL and complicates the design of methods tailored for asynchronous settings \cite{Async, AsyncOnline, AsyncTier, Async2, CommAsync, Asyncstuf, AsyncCommSched}.

Energy efficiency is an essential aspect of machine learning algorithms and one of the original motivations for FL \cite{SaveThePlanet}. The communication of high-dimensional models can be an energy-onerous task that clients may not be able to perform at every iteration. For this reason, it is crucial to cut the communication cost for clients \cite{Energy, FedAVG}. Further, such reduction can facilitate more frequent participation of resource-constrained devices, or stragglers, in the learning process. In addition, in asynchronous settings, where power and communication are restricted, ensuring communication efficiency reduces the risk of bottlenecks in the communication channels or power-related shutdown of clients due to excessive resource usage.

We can find a considerable amount of research in the literature on communication-efficient FL \cite{Energy, FedAVG, Compression, Vinay, DblCompress, RandomMask,FLCompress1,FLCompress2,FLCompress3, CommFL} and asynchronous FL \cite{Async, AsyncOnline, AsyncTier, Async2, CommAsync, Asyncstuf, TIFL, D2D, SAFA_async, AsyncCommSched}; however, only a few works consider both aspects within the same framework. The classical federated averaging (FedAvg) \cite{FedAVG}, developed for ideal conditions, reduces the communication cost by selecting a subset of the clients to participate at each iteration. In a perfect setting, this allows clients to space out their participation while maintaining a consistent participation rate. In the asynchronous settings, however, clients may already participate sporadically because of their inherent limitations. Hence subsampling comes with an increased risk of discarding valuable information. The works in \cite{AsyncTier,DblCompress} reduce communication overhead in uplink via compressed client updates. Aside from the accuracy penalty associated with the sparsification and projection used, the resulting extra computational burden on the clients of these non-trivial operations is not appealing for resource-constrained clients. Moreover, the work in \cite{DblCompress} did not consider asynchronous settings. Although the work in  \cite{AsyncTier} considers various participation frequencies for the clients, it assumes they are constant throughout the learning process. The work in \cite{CommAsync}  reduces the communication load of clients in asynchronous settings; however, it is specific to deep neural networks and lacks mathematical analysis. In addition, the considered asynchronous settings do not include communication delays. We note that structure and sketch update methods suffer the same accuracy cost and additional computational burden as compressed updates; and in all three, the simultaneous unpacking of all the received updates at the server can form a computational bottleneck. Another option explored recently for distributed learning is the partial-sharing of model parameters  \cite{PartialSharing}. The partial-sharing-based online FL (PSO-Fed) algorithm \cite{Vinay} features reduced communications in FL, but only in ideal settings.

This paper proposes a partial-sharing-based asynchronous online federated learning (PAO-Fed) algorithm for nonlinear regression in asynchronous settings. The proposed approach requires minimal communication overhead while retaining fast convergence. In order to perform nonlinear regression, we use random Fourier feature space (RFF) \cite{RFFKLMS,RFFKLMSB}, where inner products in a fixed-dimensional space approximate the nonlinear relationship between the input and output data. Consequently, given the constant communication and computational load, RFF is more suitable for decentralized learning than traditional dictionary-based solutions whose model order depends on the sample size. In addition, RFF presents the advantage of being resilient to model change during the learning process, which is key in online FL. Further, we implement partial-sharing-based communications to reduce the communication load of the algorithm. Compared to the other available methods, partial-sharing does not incur an additional computational load and only transfers a fraction of the model parameters between clients and the server. This allows clients to participate more frequently while maintaining minimal communication overhead and without additional computational burden. The proposed aggregation mechanism handles delayed updates and calibrates their contribution to the global shared model. We provide first- and second-order convergence analyses of the PAO-Fed algorithm in a setting where client participation is random and communication links suffer delays. Finally, we propose simulations on synthetic and real-life data to study the proposed algorithm and compare it with existing methods.

The paper is organized as follows. Section II introduces FL for nonlinear regression as well as partial-sharing-based communications. Section III defines the considered asynchronous settings and introduces the proposed method. Section IV provides the first and second-order convergence analysis of the PAO-Fed algorithm. Section V presents numerical results for the proposed method and compares it with existing ones. Finally, Section VI concludes the paper.

\section{Preliminaries and Problem Formulation}

This section presents the nonlinear regression problem in the context of FL. Further, a brief overview of the most closely related existing algorithms is proposed. Finally, the behavior of partial-sharing-based communications is presented.

\subsection{Online Federated Learning for Nonlinear Regression}

We consider a server connected to a set $\mathcal{K}$ of $|\mathcal{K}| = K$ geographically distributed devices, referred to as clients. In the online FL setting \cite{OnlineOrigin}, used when real-time computation is desirable, the entire dataset of a client is not immediately available. Instead, it is made available to the client progressively throughout the learning process. We denote the continuous streaming data appearing at client $k \in \mathcal{K}$ at iteration $n$ by ${\bf x}_{k,n} \in \mathbb{R}^L$, the corresponding output $y_{k,n}$ is given by:
\begin{align}
    y_{k,n} = f({\bf x}_{k,n}) + \eta_{k,n},
\end{align}
where $f(\cdot) : \mathbb{R}^L \longrightarrow \mathbb{R}$ is a nonlinear model and $\eta_{k,n}$ is the observation noise. The objective is that server and clients learn a global shared nonlinear model from the data available at each client, without this data being shared amongst clients or with the server. To this aim, the clients periodically share with the server their local model, learned from local data, and the server shares its global model with the clients.

Several adaptive methods can be used to handle nonlinear model estimation problems, e.g., \cite{KLMS,GKLMS,RFFKLMSB,RFFKLMS}. The conventional kernel least-mean-square (KLMS) algorithm \cite{KLMS} is one of the most popular choices but suffers from a growing dimensionality problem leading to prohibitive computation and communication requirements. Coherence-check-based methods \cite{GKLMS} sparsify the original dictionary by selecting the regressors using a coherence measure. Although feasible, this method is not attractive for online FL, especially in asynchronous settings, since it requires that each new dictionary element be made available throughout the network, inducing a significant communication overhead, especially if the underlying model changes. The random Fourier feature (RFF) space method \cite{RFFKLMS,RFFKLMSB} approximates the kernel function evaluation by projecting the model into a pre-selected fixed-dimensional space. The selected RFF-space does not change throughout the computation, and, given that the chosen dimension is large enough, the obtained linearizations can be as precise as desired. Therefore, we use RFF-based KLMS for the nonlinear regression task, as it is data-independent, resilient to model change, and does not require extra communication overhead, unlike conventional or coherence-check-based KLMS.

In the following, we approximate the nonlinear model by projecting it on a $D$-dimensional RFF-space, in which the function $f(\cdot)$ is approximated by the linear model  ${\bf w}^*$. To estimate the global shared model using the local streaming data, we solve the following problem:
\begin{align}
    \min_{{\bf w}}  \mathcal{J}({\bf w}),
\end{align}
where $\mathcal{J}({\bf w})$ is given by:
\begin{align}
    \mathcal{J}({\bf w}) &=  \frac{1}{K} \sum_{k \in \mathcal{K}} \mathcal{J}_k({\bf w}) \\
    \mathcal{J}_k({\bf w} )&= \mathbb{E} [|y_{k,n} - {\bf w}^{\mathsf{T}} {\bf z}_{k,n} |^2 ], \notag
\end{align}
\noindent and ${\bf z}_{k,n}$ is the mapping of ${\bf x}_{k,n}$ into the RFF-space.

\subsection{Existing Algorithms}

The Online-Fed algorithm, an online FL version of the conventional FedAvg algorithm \cite{FedAVG} solves the above estimation problem as follows. At each iteration $n$, the server selects a subset of the clients $\mathcal{K}_n \subseteq \mathcal{K}$ to participate in the learning task and shares the global shared model ${\bf w}_{n}$ with them. Then the selected clients in $\mathcal{K}_n$ perform the local learning process on their local estimates ${\bf w}_{k,n}$ as 
\begin{align}
   {\bf w}_{k,n+1} = {\bf w}_{n} + \mu {\bf z}_{k,n} e_{k,n},
\end{align}
\noindent where $\mu$ is the learning rate and $e_{k,n}$ is the \emph{a priori} error of the global model on the local data given by
\begin{align}
   e_{k,n} = y_{k,n} - {\bf w}_{n}^{\mathsf{T}} {\bf z}_{k,n}.
\end{align}
The clients then share their updated models with the server, who aggregates them as
\begin{align}
    \label{Aggregation_basic}
    {\bf w}_{n+1} = \frac{1}{| \mathcal{K}_n |} \sum_{k \in \mathcal{K}_n} {\bf w}_{k,n+1},
\end{align}
where $| \mathcal{K}_n |$ denotes the cardinality of $\mathcal{K}_n$. In the particular case where $\forall n, \mathcal{K}_n = \mathcal{K}$, i.e., all the clients participate at each iteration, we denote the algorithm Online-FedSGD.

The PSO-Fed algorithm proposed in \cite{Vinay} uses partial-sharing-based communications to reduce further the communication overhead of the Online-Fed algorithm. Additionally, PSO-Fed allows clients that are not participating in the current iteration to perform local learning on their new data. Doing so, this algorithm drastically reduces the communication overhead without compromising the convergence speed.

\subsection{Partial-sharing-based Communications}

\begin{figure}
    \centering
    \includegraphics[scale=0.45]{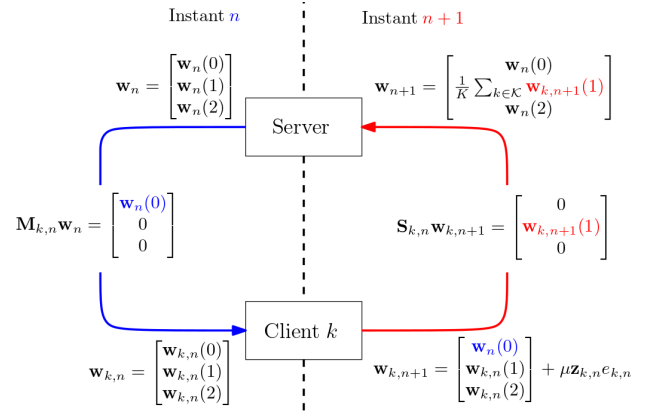}
    \caption{Partial sharing in a simple scenario.}
    \label{PS}
\end{figure}

In partial-sharing-based communications, as defined in \cite{PartialSharing}, the server and the clients exchange only a portion of their respective models instead of the entire model. The portion is extracted prior to communication by multiplication with a diagonal selection matrix with main diagonal elements being either $0$ or $1$, where the locations of the latter specify the model parameters to share. This operation is computationally trivial and, therefore, does not induce delay on the communication, unlike compressed update methods, e.g., \cite{AsyncTier,DblCompress}. $m$ denotes the number of nonzero elements in the selection matrices, this is the number of model parameters shared at each iteration. The selection matrix ${\bf M}_{k,n}$ is used for server-to-client communication at time $n$ and the selection matrix ${\bf S}_{k,n}$ for client $k$'s response, as can be seen on Fig. \ref{PS} where the simple case where $m = D/3$ is illustrated.

The usual aggregation step in \eqref{Aggregation_basic} cannot be used with partial-sharing-based communications, it has to be adapted. The expression of ${\bf w}_{n+1}$ in Fig. \ref{PS} is the aggregation step for coordinated partial-sharing in perfect settings. Coordinated partial-sharing is the special case where all clients send the same portion of the model at a given iteration.

For the clients to participate in the learning of the whole model, and to ensure consistency across models, it is necessary that the selection matrices evolve. To this aim, we set:
\begin{align}
    \label{SelectionMatrixChoice}
    \text{diag}({\bf M}_{k,n+1}) &= \text{circshift}(\text{diag}({\bf M}_{k,n}),m) \\
    \label{SelectionMatrixChoice2}
    {\bf S}_{k,n} &= {\bf M}_{k,n+1}
\end{align}
where $\text{circshift}$ denotes a circular shift operator. ${\bf S}_{k,n}$ is set to be equal to ${\bf M}_{k,n+1}$ rather than ${\bf M}_{k,n}$ in order to share a portion of the client's model further refined by the local learning process. As can be seen in Fig. \ref{PS}, ${\bf w}_{k,n+1}(0)$ contains information from a single local learning step of client $k$, while ${\bf w}_{k,n+1}(1)$ contains information from three (since it is equal to ${\bf w}_{n-2}(1)$ refined thrice by the local learning process). For smaller values of $m$, the additional information is greater, but the original value of the portion is also older.

\subsection{Motivation}

The above mentioned algorithms offer significant communication overhead cuts; however, they do not consider practical network environment and client resources. When performing federated learning in real-world applications, clients may be unavailable for various reasons, message exchanges may be delayed or blocked, and straggler clients may be present. For this reason, it is essential to tailor the developed algorithms to asynchronous settings. Those environments impact the algorithm design and optimization. For instance, we will see that many choices made for the PSO-Fed algorithm in an ideal setting are unsuitable for asynchronous settings.

\section{Proposed Method}

This section presents the proposed communication-efficient Partial-sharing-based Asynchronous Online Federated Learning (PAO-Fed) algorithm and the asynchronous settings for which it is developed. 
\subsection{Asynchronous Settings}

The following features are necessary to have for an online FL method to operate successfully in realistic environments.
\begin{itemize}
    \item The capability to handle non-IID and unevenly distributed data.

    \item The capability to handle heterogeneous, time-varying, and unpredictable client participation, including possible downtimes. In most real-world applications, the computation and communication capacity of a specific task are heterogeneous and time-varying, in addition, clients are unreliable as they may experience many issues (low battery, software failure, physical threat, etc.). Moreover, when dealing with many clients, an infrequently occurring failure is likely experienced at least once. Lastly, it is unlikely for the server to know in advance when a client will be unavailable or suffer a failure, even the most reliable clients may suffer downtimes.
    
    \item The capability to weigh the importance of delayed messages. Model parameters with the same timestamp may arrive at different instants at the server. In practice, communication channels are unreliable and although most messages arrive within a short window, some may take longer, especially when the communication channels are strained. In addition, straggler clients may not be able to complete the learning task in the given time-frame, and although their update may not be delayed, it will arrive late at the server. Therefore, the developed method must be robust to a delay spread in the received parameters.
    
    \item The capability to reduce the likelihood of straggler-like behavior. Resource-constrained devices may induce latency or run out of power, resulting in reduced information sharing. It is, therefore, not sufficient to consider stragglers-like behavior \cite{AsyncOnline}; it is preferable to improve their operational environments, e.g., by reducing their computation and communication load.
\end{itemize}

The first step to address those challenges is to properly model the presented behaviors. To this aim, the clients participation is modeled by participation probabilities. At an iteration $n$, the Bernoulli trial on the probability $p_{k,n}$ dictates if client $k$ is able to participate. The use of probabilities for participation allows the model to address all the behaviors presented in the second point, unlike the commonly used tier-based model for participation (e.g., \cite{AsyncTier}), where each tier is expected to behave optimally given a tailored frequency. In fact, heterogeneity and time-dependency is handled by giving clients various evolving probabilities $p_{k,n}$, and unpredictability and downtimes are naturally present when ensuring that all probabilities are lower than one. In addition, any communication sent by a client to the server may be delayed by one or several iterations.

With the proposed model, the limitations of real-world applications and the heterogeneity of the computational power and communication capacity of the available devices is taken into consideration. Those asynchronous settings diminish the potential performance of an FL method, especially in the online setting, where data not shared in time is lost. The proposed method ensures communication efficiency and, in turn, some extend energy-efficiency in order, notably, to avoid downward cycles in the asynchronous behavior of the participating clients. For instance, a weaker device may take longer to perform the learning process, struggle to send a long message, and need time to save enough power to participate again. Therefore, performing less computation and exchanging shorter messages will reduce the burden on the clients and the communication channels, making further complications or delays less likely. For this reason, a communication-efficient method tailored for the asynchronous settings can perform above its expectations in a real-life scenario.

\subsection{Delayed Updates}

The consequence of the introduced delays is that not all updates sent by clients participating at a given iteration will arrive at the server simultaneously. Precisely, we denote $\mathcal{K}_n$ the set of all the clients who sent an update that arrived at the server at iteration $n$. This set can be decomposed as:
\begin{equation}
    \mathcal{K}_n = \bigcup_{l=0}^{\infty} \mathcal{K}_{n,l}
\end{equation}
\noindent where $\mathcal{K}_{n,l}$ denotes the set of the clients who sent an update at iteration $n-l$ which reached the server at iteration $n$, the subscript $l$ corresponds to the number of iterations during which the update was delayed. A delayed update will naturally lose value the longer it is delayed, as it becomes outdated. To improve the learning accuracy of the proposed algorithm, we propose a weight-decreasing mechanism that weights down updates that have been delayed. By doing so, we diminish the negative impact of outdated data on the convergence. This mechanism is different from age of update mechanisms found in \cite{AoI, AoU, AoU2} where weights are dictated by the amount of data, independently from communication delays. We denote $\alpha_l \in [0,1]$ the weight given to the updates sent by the clients in $\mathcal{K}_{n,l}$. In this work, we only consider delays in clients-to-server communications. Although delays in server-to-clients communications diminish performance as well, they do not require further modification of the aggregation mechanism as they can be handled in the same manner. For simplicity reason, we omit those in the following.

\subsection{PAO-Fed}

The proposed PAO-Fed algorithm is tailored to the asynchronous settings, notably, its novel aggregation step is designed to handle delayed updates. PAO-Fed makes use of all the available clients at a given iteration. To reduce the communication overhead associated with the learning, it uses partial-sharing-based communications, which is well adapted to the asynchronous settings as it does not lay any additional computational burden on the participating clients. Further, the aggregation step is refined with a weight-decreasing mechanism to diminish the negative impact of delayed updates on convergence. The algorithm is as follows.

During iteration $n$, the server shares a portion of the global shared model, i.e. ${\bf M}_{k,n} {\bf w}_n$, to all the available clients. The selection matrix ${\bf M}_{k,n}$ dictates which portion of the model is sent to client $k$. The available client $k$ receives its portion of the global shared model, and uses it to update its local model, the new local model is given by ${\bf M}_{k,n} {\bf w}_n + ({\bf I} - {\bf M}_{k,n}) {\bf w}_{k,n}$. Afterwards, the available client $k$ refines its local model by performing the process of local learning on its newly available data as follows:
\begin{align}
\label{ClientUpdateSelected}
    {\bf w}_{k,n+1} &= {\bf M}_{k,n} {\bf w}_n + ({\bf I} - {\bf M}_{k,n}) {\bf w}_{k,n} + \mu {\bf z}_{k,n} e_{k,n}, 
\end{align}
where $e_{k,n}$ is the \emph{a priori} error of the local model on the local data given by:
\begin{align}
\label{Error_available}
    e_{k,n} &=  y_{k,n} - ({\bf M}_{k,n} {\bf w}_n + ({ \bf I} - {\bf M}_{k,n}) {\bf w}_{k,n} )^{\text{T}} {\bf z}_{k,n}.
\end{align}

When a client is unavailable at a given iteration but receives new data and is not malfunctioning, it refines its local model autonomously. For example, this can be a case where a client is well functioning but do not have communication capacity at the time. This local update step, identical to the one used in \cite{Vinay}, is performed as:
\begin{align}
    \label{LocalUpdateRe}
   {\bf w}_{k,n+1} = {\bf w}_{k,n} + \mu {\bf z}_{k,n} e_{k,n},
\end{align}
\noindent where $e_{k,n}$ in that case is given by:
\begin{align}
    e_{k,n} = y_{k,n} - {\bf w}_{k,n}^{\mathsf{T}} {\bf z}_{k,n}.
\end{align}
\noindent This update is computationally trivial for most devices and does not involve communication. Its purpose is for the client to share better refined model parameters during the next participation. Naturally, this additional information only reaches the server if the model parameters are not overwritten before being communicated, further motivating the choice made in \eqref{SelectionMatrixChoice2}.

After this local update step, all available clients communicate a portion of their updated local models to the server. A client $k$ communicates the portion of the model dictated by the selection matrix ${\bf S}_{k,n}$, that is, ${\bf S}_{k,n} {\bf w}_{k,n+1}$. Those updates may arrive at the present iteration or at a later one if they are delayed.

At the server, we consider the previously introduced set $\mathcal{K}_n$ consisting of the clients whose updates arrive at the current iteration. This set comprises of the sets $\mathcal{K}_{n,l}, 0 \leqslant l < \infty$ consisting of the clients whose update was sent at iteration $n-l$ and arrive at the current iteration. The set $\mathcal{K}_{n,0}$ consists of the available clients at the current iteration whose updates have not been delayed. Note that a client may appear twice in the set $\mathcal{K}_n$ if two of its updates arrive at the same iteration. The deviation from the current global model engendered by the updates received from a non-empty set $\mathcal{K}_{n,l}$ can be expressed as:
\begin{align}
\label{Deviation}
   \Delta_{n,l} &= \frac{1}{|\mathcal{K}_{n,l}|} \sum_{k \in \mathcal{K}_{n,l}} {\bf S}_{k,n-l} ({\bf w}_{k,n+1-l} - {\bf w}_n).
\end{align}
If a set $\mathcal{K}_{n,l}$ is empty, we set by convention $\Delta_{n,l} = {\bf 0}$

The aggregation step of the proposed algorithm uses a weight-decreasing mechanism for the delayed updates. A client's participation that have been delayed for $l$ iterations will be given the weight $\alpha_l \in [0,1]$. By convention, we set the weight of the updates that are not delayed to $\alpha_0 = 1$. The resulting aggregation mechanism is given by:
\begin{align}
\label{ServerUpdate}
    {\bf w}_{n+1} &= {\bf w}_n + \sum_{l = 0}^{\infty} \alpha_l \Delta_{n,l} .
\end{align}
When $l > l_{\text{max}}$, the maximum effective delay, the aggregation mechanism discards the corresponding updates by setting $\alpha_l = 0, \: l > l_{\text{max}}$. It is possible to replace $\infty$ by $l_{\text{max}}$ in \eqref{ServerUpdate} without changing the aggregation mechanism. Note that in the eventuality where several updates from clients in $\mathcal{K}_{n}$ update the same model parameter, only the most recent updates are considered, the selection matrices of the remaining updates are adjusted accordingly prior to computing \eqref{ServerUpdate}. The resulting algorithm is presented in Algorithm \ref{PAO-Fed}.

\begin{algorithm}[t!]
\caption{PAO-Fed}
\label{PAO-Fed}
\begin{algorithmic}[1]
    \STATE Initialization: ${\bf w}_0$ and ${\bf w}_{k,0}, k \in \mathcal{K}$ set to ${\bf 0}$
    \STATE Procedure at Local client $k$
    \FOR{ iteration $n=1,2,\hdots,N$}
        \IF{Client $k$ receives new data at time $n$}
            \IF{$k$ is available}
                \STATE Receive ${\bf M}_{k,n} {\bf w}_{n}$ from the server.
                \STATE Compute ${\bf w}_{k,n+1}$ as in \eqref{ClientUpdateSelected}.
                \STATE Share ${\bf S}_{k,n} {\bf w}_{k,n+1}$ with the server.
            \ELSE
                \STATE Update ${\bf w}_k$ as in \eqref{LocalUpdateRe}.
            \ENDIF
        \ENDIF
    \ENDFOR
    \STATE Procedure at Central Server
    \FOR{ iteration $n=1,2,\hdots,N$}
        \STATE Receive client updates from subset $\mathcal{K}_n \subset \mathcal{K}$.
        \STATE Compute ${\bf w}_{n+1}$ as in \eqref{ServerUpdate}.
        \STATE Share ${\bf M}_{k,n+1} {\bf w}_{n+1}$ with the available clients. 
	\ENDFOR
\end{algorithmic} 
\end{algorithm}

\subsection{Partial-sharing in Asynchronous Settings}

In coordinated partial sharing, all participating clients share the same portion of the model so that the server's model is aggregated from a large number of clients, thus improving accuracy. For this reason, coordinated partial-sharing is used in most algorithms assuming perfect settings. In practice, however, delayed updates partially overwrite the previously aggregated portion, as can be seen in \eqref{ServerUpdate}, thus negating the added value of coordination.

To tackle this issue, one can either use a weight-decreasing mechanism such as the one presented above, or use uncoordinated partial sharing. Besides, uncoordinated partial-sharing is ideal when dealing with underlying model changes, as the server's model uniformly steer towards its new steady-state value, instead of doing so portion by portion as with coordinated partial-sharing.

\section{Convergence Analysis}
In this section, we examine the convergence behavior of the proposed PAO-Fed algorithm that uses partial-sharing-based communications and evolves in asynchronous settings such as the ones presented in Section III. We prove mathematically that the proposed PAO-Fed algorithm converges to the exact model in the RFF space and exhibits stable extended mean square displacement under certain general assumptions.

Before proceeding to the analysis, we introduce auxiliary matrices to express an entire iteration of the algorithm in the matrix form. Similar to \cite{tcas2}, we define the extended model vector ${\bf w}_{e,n}$, local update matrix ${\bf A}_{e,n}$, and mapping of the data into the RFF-space ${\bf Z}_{e,n}$ as
\begin{align}
    {\bf w}_{e,n} &= \text{col} \{ {\bf w}_{n},{\bf w}_{1,n}, \ldots, {\bf w}_{K,n}, {\bf w}_{1,n} \ldots, {\bf w}_{K,n}, {\bf w}_{1,n-1}, \notag\\
    & \hspace{12mm} \ldots, {\bf w}_{K,n-1}, \ldots, {\bf w}_{1,n-l_{\text{max}}}, \ldots, {\bf w}_{K,n-l_{\text{max}}} \}, \notag\\
    {\bf A}_{e,n} & = \text{blockdiag} \{ {\bf A}_n , {\bf I}_{DK}, \ldots, {\bf I}_{DK}\}, \notag\\
    {\bf Z}_{e,n} &= \text{blockdiag} \{{\bf Z}_{n}, {\bf 0}_{DK \times K}, \ldots, {\bf 0}_{DK \times K} \}, 
\end{align}
with 
\begin{align}
    {\bf A}_n &= 
    \begin{bmatrix}
     {\bf I} & {\bf 0}_{D} & \dotsb & {\bf 0}_{D} \\
     a_{1,n} {\bf M}_{1,n} & {\bf I} - a_{1,n} {\bf M}_{1,n} &  & \vdots \\
     \vdots & {\bf 0}_{D} & \ddots & {\bf 0}_{D} \\
     a_{K,n} {\bf M}_{K,n} & \vdots & & {\bf I} - a_{K,n} {\bf M}_{K,n} \\
     \end{bmatrix},
     \notag\\
     {\bf Z}_{n} &= \text{blockdiag}\{{\bf 0}_{D}, {\bf z}_{1,n}, \ldots, {\bf z}_{K,n}\} ,
\end{align}
where $a_{k,n} = 1$ if the client $k$ is available at iteration $n$ and $0$ otherwise,  $\text{col}\{\cdot\}$ and $\text{blockdiag}\{\cdot\}$ represent column-wise stacking and block diagonalization operators, respectively. We can now express the extended observation vector ${\bf y}_{e,n} = \text{col}\{0,   y_{1, n}, y_{2, n}, \ldots, y_{K, n}, {\bf 0}_{K \times 1}, \ldots, {\bf 0}_{K \times 1}\}$ as
\begin{align}
    {\bf y}_{e,n} = {\bf Z}_{e,n}^{\mathsf{T}} {\bf w}_{e}^* + \boldsymbol{\eta}_{e,n},
\end{align}
where ${\bf w}_{e}^* = {\bf 1}_{(K+1)l_{\text{max}} +1} \otimes {\bf w}^*$ and the extended observation noise $\boldsymbol{\eta}_{e, n}  = \text{col}\{0, \eta_{1, n},  \eta_{2, n}, \ldots,\eta_{K, n}, {\bf 0}_{K \times 1}, \ldots, {\bf 0}_{K \times 1}\}$. We then can express the extended estimation error vector as
\begin{align}
    {\bf e}_{e,n} = {\bf y}_{e,n} - {\bf Z}_{e,n}^\mathsf{T} {\bf A}_{e,n} {\bf w}_{e,n}.
\end{align}

Therefore, the recursion of the extended model vector ${\bf w}_{e,n}$ is given by
\begin{align}
\label{Total_update_bigmatrices}
    {\bf w}_{e,n+1} = {\bf B}_{e,n} ({\bf A}_{e,n} {\bf w}_{e,n} + \mu {\bf Z}_{e,n} {\bf e}_{e,n}),
\end{align}
with
\begin{align}
    {\bf B}_{e,n} &=
    \begin{bmatrix}
     {\bf B}_{n} & {\bf B}_{0,n} & {\bf 0}_{D \times DK}  & {\bf B}_{1,n} & \dotsb & {\bf B}_{l_{\text{max}},n} \\
     {\bf 0}_{D \times 1} & {\bf I}_{D K} & {\bf 0}_{D K} & \dotsb & \dotsb & {\bf 0}_{D K} \\
     \vdots & {\bf I}_{D K} & {\bf 0}_{D K} & \dotsb & \dotsb & {\bf 0}_{D K} \\
     \vdots & {\bf 0}_{D K} & {\bf I}_{D K} & {\bf 0}_{D K} & \dotsb & {\bf 0}_{D K} \\
     \vdots & \vdots & \ddots & \ddots & \ddots & {\bf 0}_{D K}\\
     {\bf 0}_{D \times 1} & {\bf 0}_{D K} & \dotsb & {\bf 0}_{D K} & {\bf I}_{D K} & {\bf 0}_{D K} \\
     \end{bmatrix}
     \notag \\
     {\bf B}_{n} &= {\bf I} - \sum_{l=0}^{l_{\text{max}}} \alpha_l \sum_{k \in \mathcal{K}_{n,l}} \frac{b_{k,n,l} }{|\mathcal{K}_{n,l}|} {\bf S}_{k,n-l} \notag \\
     {\bf B}_{l,n} &= [\frac{\alpha_l b_{1,n,l}}{|\mathcal{K}_{n,l}|} {\bf S}_{1,n-l}, \dotsb, \frac{\alpha_l b_{K,n,1}}{|\mathcal{K}_{n,l}|} {\bf S}_{K,n-l}].
\end{align}
\noindent where $b_{k,n,l} = 1$ if $k \in \mathcal{K}_{n,l}$ and $0$ otherwise.

In the following, we present detailed convergence analysis of the PAO-Fed algorithm both in mean and mean-square senses. To this end, we make the following assumptions:

\noindent \textbf{Assumption 1:} The mapped data vectors ${\bf z}_{k,n}$ are drawn at each time step from a WSS multivariate random sequence with correlation matrix ${\bf R}_k = \mathbb{E}[{\bf z}_{k,n} {\bf z}^\mathsf{T}_{k,n}]$.

\noindent \textbf{Assumption 2:} The observation noise $\eta_{k,n}$ is assumed to be zero mean white Gaussian, and independent of all input and output data.

\noindent \textbf{Assumption 3:} At each client, the model parameter vector is assumed to be independent of the input data.

\noindent \textbf{Assumption 4:} The selection matrices are assumed independent from each other, and of any other data.

\noindent \textbf{Assumption 5:} The learning rate $\mu$ is small enough for terms involving higher-order powers of $\mu$ to be neglected.

\noindent \textit{Note:} No assumption on the $\alpha_{l}$ variables is required as long as $l_{\text{max}}$ is defined.

\subsection{First-order Analysis}

This subsection examines the mean convergence of the proposed PAO-Fed algorithm.

\begin{theorem}
Let \textbf{Assumptions 1--4} hold true. Then, The proposed PAO-Fed converges in mean if and only if
\begin{align}
\label{Convergence_condition}
    0 < \mu < \frac{2}{ \max\limits_{\forall k, i} \lambda_i({\bf R}_k) }.
\end{align}
\end{theorem} 

\begin{IEEEproof}
Denoting the model error vector $\Tilde{{\bf w}}_{e,n} = {\bf w}_e^* - {\bf w}_{e,n}$, and using the fact that ${\bf w}_e^* = {\bf B}_{e,n} {\bf A}_{e,n}  {\bf w}_e^*$ (by construction, all rows in ${\bf B}_{e,n}$ and ${\bf A}_{e,n}$ sum to $1$), from \eqref{Total_update_bigmatrices}, we can recursively express $\Tilde{{\bf w}}_{e,n}$ as
\begin{align}
    \label{error_recursion}
    \Tilde{{\bf w}}_{e,n+1} &= {\bf w}_e^* - {\bf w}_{e,n+1} \notag\\
    &= {\bf w}_e^* - {\bf B}_{e,n} {\bf A}_{e,n} {\bf w}_{e,n} - {\bf B}_{e,n} \mu {\bf Z}_{e,n} {\bf e}_{e,n} \notag\\
    &= {\bf B}_{e,n} {\bf A}_{e,n} \Tilde{{\bf w}}_{e,n} - {\bf B}_{e,n} \mu {\bf Z}_{e,n} \boldsymbol{\eta}_{e,n} \notag \\
    & \hspace{5mm}- {\bf B}_{e,n} \mu {\bf Z}_{e,n} {\bf Z}_{e,n}^\mathsf{T} ({\bf w}_{e}^* - {\bf A}_{e,n} {\bf w}_{e,n}) \notag \\
    &= {\bf B}_{e,n} ({\bf I} - \mu {\bf Z}_{e,n} {\bf Z}_{e,n}^\mathsf{T}) {\bf A}_{e,n} \Tilde{{\bf w}}_{e,n} \notag \\
    &\hspace{5mm} - \mu {\bf B}_{e,n} {\bf Z}_{e,n} \boldsymbol{\eta}_{e,n}. 
\end{align}
Taking the statistical expectation $\mathbb{E} [\cdot] $ on both sides of \eqref{error_recursion} and using \textbf{Assumptions $1$--$4$}, we obtain
\begin{align}
    \label{1stOrder}
    \mathbb{E} [\Tilde{{\bf w}}_{e,n+1}] &= \mathbb{E} [{\bf B}_{e,n}] \mathbb{E} [{\bf I} - \mu {\bf Z}_{e,n} {\bf Z}_{e,n}^\mathsf{T}] \mathbb{E} [{\bf A}_{e,n}] \mathbb{E} [\Tilde{{\bf w}}_{e,n}] \notag\\
    &= \mathbb{E} [{\bf B}_{e,n}] ({\bf I} - \mu {\bf R}_e) \mathbb{E} [{\bf A}_{e,n}] \mathbb{E} [\Tilde{{\bf w}}_{e,n}],
\end{align}
where ${\bf R}_e = \text{blockdiag}\{ {\bf 0}_{D}, {\bf R}_1, {\bf R}_1, \dotsb, {\bf R}_K, {\bf 0}_{DK l_{\text{max}}} \}$. The quantities $\mathbb{E} [{\bf A}_{e,n}]$ and $\mathbb{E} [{\bf B}_{e,n}]$ are evaluated in Appendix A.

Further, we consider the vectors and matrices reduced to the subspace between the index $D+1$ and $D(K+1)$. We denote the reduction of ${\bf x}$ by ${\bf x} |_{\text{sel}}$. Using the reduced definitions, \eqref{1stOrder} becomes: $ \mathbb{E} [\Tilde{{\bf w}}_{e,n+1}|_{\text{sel}}] = ({\bf I} - \mu {\bf R}_{e}|_{\text{sel}}) \mathbb{E} [{\bf A}_{e,n}|_{\text{sel}}] \mathbb{E} [\Tilde{{\bf w}}_{e,n}|_{\text{sel}}]$, where the block $ \Tilde{{\bf w}}_{e,n}|_{\text{sel}}$ is defined as a linear sequence of order $1$ in a normed algebra. To prove the convergence of $\mathbb{E} [\Tilde{{\bf w}}_{e,n}|_{\text{sel}}]$, we use the properties of the block maximum norm \cite{BlockMax}. From Appendix A, we have $|| \mathbb{E} [{\bf A}_{e,n}|_{\text{sel}}] ||_{b,\infty} = 1$. Then the convergence condition reduces to $||{\bf I} - \mu {\bf R}_{e}|_{\text{sel}} ||_{b,\infty} < 1$, equivalently,  $|1 - \mu \lambda_i({\bf R}_k)| < 1$, $\forall k, i$, where $\lambda_i(\cdot)$ is the $i$th eigenvalue of the argument matrix. This leads to the convergence condition given by \eqref{Convergence_condition}.

\end{IEEEproof}

\subsection{Second-order Analysis}

In this subsection, we present the second-order analysis of the proposed PAO-Fed algorithm. For the given arbitrary positive semidefinite matrix ${\bf \Sigma}$, the weighted norm-square of $\Tilde{{\bf w}}_{e,n}$ is given by $|| \Tilde{{\bf w}}_{e,n} ||_{{\bf \Sigma}}^2 = \Tilde{{\bf w}}_{e,n}^\mathsf{T} {\bf \Sigma} \Tilde{{\bf w}}_{e,n}$. From \eqref{error_recursion}, we can obtain
\begin{align}
\label{exprecursion}
    \mathbb{E} [ || \Tilde{{\bf w}}_{e,n+1} ||_{{\bf \Sigma}}^2 ] = \mathbb{E} [ || \Tilde{{\bf w}}_{e,n} ||_{{\bf \Sigma}'}^2 ] + \mu^2 \mathbb{E} [ \boldsymbol{\eta}_{e,n}^{\mathsf{T}} {\bf Y}_n^{{\bf \Sigma}} \boldsymbol{\eta}_{e,n} ],
\end{align}
where the cross terms are null under \textbf{Assumption 2} and the matrices ${\bf \Sigma}'$ and ${\bf Y}^{{\bf \Sigma}}$ are given by
\begin{align}
    {\bf \Sigma}' &= \mathbb{E} [ {\bf A}_{e,n}^{\mathsf{T}} ({\bf I} - \mu {\bf Z}_{e,n} {\bf Z}_{e,n}^\mathsf{T}) {\bf B}_{e,n}^{\mathsf{T}} \: {\bf \Sigma} \\
    & \hspace{8mm} {\bf B}_{e,n} ({\bf I} - \mu {\bf Z}_{e,n} {\bf Z}_{e,n}^\mathsf{T}) {\bf A}_{e,n} ], \notag \\
    {\bf Y}_n^{{\bf \Sigma}} &= {\bf Z}_{e,n}^\mathsf{T} {\bf B}_{e,n}^{\mathsf{T}} {\bf \Sigma} {\bf B}_{e,n} {\bf Z}_{e,n}.
\end{align}

Using \textbf{Assumption 3} and the properties of the block Kronecker product,  and the block vectorization operator $\text{bvec}\{ \cdot \}$ \cite{Kro}, we can establish a relationship between $\boldsymbol{\sigma} = \text{bvec}\{ {\bf \Sigma} \}$ and $\boldsymbol{\sigma}' = \text{bvec}\{ {\bf \Sigma}' \}$ as
\begin{align}
\label{sigmalink}
    \boldsymbol{\sigma}' &= \boldsymbol{\mathcal{F}}^{\mathsf{T}} \boldsymbol{\sigma},
\end{align}
where
\begin{align}
    \boldsymbol{\mathcal{F}} &= \boldsymbol{\mathcal{Q}}_{{\bf B}} \boldsymbol{\mathcal{Q}}_{{\bf A}} - \mu \boldsymbol{\mathcal{Q}}_{{\bf B}} ({\bf I} \otimes_b {\bf R}_e) \boldsymbol{\mathcal{Q}}_{{\bf A}} - \mu \boldsymbol{\mathcal{Q}}_{{\bf B}} ({\bf R}_e \otimes_b {\bf I}) \boldsymbol{\mathcal{Q}}_{{\bf A}}, \notag
\end{align}
where the higher-order powers of $\mu$ are neglected under \textbf{Assumption 5}. In the above
\begin{align}
\label{Qmatrices}
    \boldsymbol{\mathcal{Q}}_{{\bf A}} &= \mathbb{E} [ {\bf A}_{e,n} \otimes_b {\bf A}_{e,n} ], \notag\\
    \boldsymbol{\mathcal{Q}}_{{\bf B}} &= \mathbb{E} [ {\bf B}_{e,n} \otimes_b {\bf B}_{e,n} ].
\end{align}

In Appendix B, we evaluate the matrices $\boldsymbol{\mathcal{Q}}_{{\bf A}}$ and $\boldsymbol{\mathcal{Q}}_{{\bf B}}$, and prove that all their entries are real, non-negative, and add up to unity on each rows. This implies that both matrices are right-stochastic, and thus, their spectral radius is equal to one.

We will now evaluate the term $\mathbb{E} [ \boldsymbol{\eta}_{e,n}^{\mathsf{T}} {\bf Y}_n^{{\bf \Sigma}} \boldsymbol{\eta}_{e,n} ]$ as follows:
\begin{align}
\label{expeliarmus}
    &\mathbb{E} [ \boldsymbol{\eta}_{e,n}^{\mathsf{T}} {\bf Y}_n^{{\bf \Sigma}} \boldsymbol{\eta}_{e,n} ] = \mathbb{E} [ \boldsymbol{\eta}_{e,n}^{\mathsf{T}} {\bf Z}_{e,n}^\mathsf{T} {\bf B}_{e,n}^{\mathsf{T}} {\bf \Sigma} {\bf B}_{e,n} {\bf Z}_{e,n} \boldsymbol{\eta}_{e,n} ] \notag\\
    &= \mathbb{E} [ \text{trace} ( \boldsymbol{\eta}_{e,n}^{\mathsf{T}} {\bf Z}_{e,n}^\mathsf{T} {\bf B}_{e,n}^{\mathsf{T}} {\bf \Sigma} {\bf B}_{e,n} {\bf Z}_{e,n} \boldsymbol{\eta}_{e,n} ) ] \notag \\
    &= \text{trace} ( \mathbb{E} [ {\bf B}_{e,n} {\bf Z}_{e,n} \mathbb{E} [ \boldsymbol{\eta}_{e,n}^{\mathsf{T}} \boldsymbol{\eta}_{e,n} ] {\bf Z}_{e,n}^\mathsf{T} {\bf B}_{e,n}^{\mathsf{T}} ] {\bf \Sigma} ) \notag \\
    &= \text{trace} ( \mathbb{E} [ {\bf B}_{e,n} {\bf \Phi}_n {\bf B}_{e,n}^{\mathsf{T}} ] {\bf \Sigma} ), 
\end{align}
with ${\bf \Phi}_n = {\bf Z}_{e,n} {\bf \Lambda}_{\eta} {\bf Z}_{e,n}^\mathsf{T}$, where ${\bf \Lambda}_{\eta} = \mathbb{E} [ \boldsymbol{\eta}_{e,n}^{\mathsf{T}} \boldsymbol{\eta}_{e,n} ]$ is a diagonal matrix having the noise variances of all clients on its main diagonal. Note that we use \textbf{Assumption 2} in the last line of \eqref{expeliarmus}. Finally, using the properties of block Kronecker product we have
\begin{align}
    &\text{trace} ( \mathbb{E} [ {\bf B}_{e,n} {\bf \Phi}_n {\bf B}_{e,n}^{\mathsf{T}} ] {\bf \Sigma} ) = {\bf h}^{\mathsf{T}} \boldsymbol{\sigma},
\end{align}
with 
\begin{align}
    {\bf h} &= \text{bvec} \{ \mathbb{E} [ {\bf B}_{e,n} {\bf \Phi}_n {\bf B}_{e,n}^{\mathsf{T}} ] \} \notag\\
    &= \boldsymbol{\mathcal{Q}}_{{\bf B}} \text{bvec} \{ \mathbb{E} [ {\bf \Phi}_n ] \}.
\end{align}

Combining \eqref{exprecursion}, \eqref{sigmalink}, and \eqref{expeliarmus}, we can write the recursion for the weighted extended mean square displacement of the PAO-Fed algorithm as:
\begin{align}
\label{baseTh2}
    \mathbb{E} [ || \Tilde{{\bf w}}_{e,n+1} ||_{\text{bvec}^{-1} \{ \boldsymbol{\sigma} \}}^2 ] = \mathbb{E} [ || \Tilde{{\bf w}}_{e,n} ||_{\text{bvec}^{-1} \{ \boldsymbol{\mathcal{F}}^{\mathsf{T}} \boldsymbol{\sigma} \}}^2 ] + \mu^2 {\bf h}^{\mathsf{T}} \boldsymbol{\sigma},
\end{align}
where $\text{bvec}^{-1} \{ \cdot \}$ represents the reverse operation of block vectorization.

\begin{theorem}
Let \textbf{Assumptions 1--5} hold true. Then, the PAO-Fed algorithm exhibits stable extended mean square displacement if and only if:
\begin{align}
\label{Convergence_condition2}
    0 < \mu < \frac{1}{ \max\limits_{\forall k, i} \lambda_i({\bf R}_k) }.
\end{align}
\end{theorem}

\begin{IEEEproof}
Iterating \eqref{baseTh2} backwards to $n=0$, we get
\begin{align}
    \mathbb{E} [ || \Tilde{{\bf w}}_{e,n+1} ||_{\text{bvec}^{-1} \{ \boldsymbol{\sigma} \}}^2 ] = &\mathbb{E} [ || \Tilde{{\bf w}}_{e,0} ||_{\text{bvec}^{-1} \{ (\boldsymbol{\mathcal{F}}^{\mathsf{T}})^{n+1} \boldsymbol{\sigma} \}}^2 ] \notag\\
    &+ \mu^2 {\bf h}^{\mathsf{T}} ({\bf I} + \sum_{j=1}^{n} (\boldsymbol{\mathcal{F}}^{\mathsf{T}})^j) \boldsymbol{\sigma}.
\end{align}

To prove the convergence of $\mathbb{E} [|| \Tilde{{\bf w}}_{e,n} ||_{{\bf \Sigma}}^2] = \mathbb{E} [|| \Tilde{{\bf w}}_{e,n+1} ||_{\text{bvec}^{-1} \{ \boldsymbol{\sigma} \}}^2 ]$, we need to prove that the spectral radius of $\boldsymbol{\mathcal{F}}$ is less than one, i.e. $ \rho(\boldsymbol{\mathcal{F}}) < 1 $. Using the properties of the block maximum norm \cite{BlockMax}, we have;
\begin{align}
    \rho(\boldsymbol{\mathcal{F}}) &\leqslant || \boldsymbol{\mathcal{Q}}_{{\bf B}} ( {\bf I} - \mu ({\bf I} \otimes_b {\bf R}_e) - \mu ({\bf R}_e \otimes_b {\bf I}) ) \boldsymbol{\mathcal{Q}}_{{\bf A}} ||_{b,\infty}, \notag \\
    &\leqslant || \boldsymbol{\mathcal{Q}}_{{\bf B}} ||_{b,\infty} || \boldsymbol{\mathcal{Q}}_{{\bf A}} ||_{b,\infty} \notag\\
    &\; \; \; \; \; || ( {\bf I} - \mu ({\bf I} \otimes_b {\bf R}_e) - \mu ({\bf R}_e \otimes_b {\bf I}) ) ||_{b,\infty}.
\end{align}

Since the matrices $\boldsymbol{\mathcal{Q}}_{{\bf A}}$ and $\boldsymbol{\mathcal{Q}}_{{\bf B}}$ are right stochastic, we have $|| \boldsymbol{\mathcal{Q}}_{{\bf A}} ||_{b,\infty} = || \boldsymbol{\mathcal{Q}}_{{\bf B}} ||_{b,\infty} = 1$. Therefore the condition $|| ( {\bf I} - \mu ({\bf I} \otimes_b {\bf R}_e) - \mu ({\bf R}_e \otimes_b {\bf I}) ) ||_{b,\infty} < 1$, equivalently, $|1 - \mu (\lambda_i ({\bf R}_e) + \lambda_j ({\bf R}_e))| < 1, \; \forall i,j$, is sufficient to guarantee the convergence of $|| \Tilde{{\bf w}}_{e,n} ||_{{\bf \Sigma}}^2$. This simplification leads to the convergence conditions in \eqref{Convergence_condition2}.
\end{IEEEproof}

\subsection{Transient and Steady-state Mean Square Deviation}

From \eqref{baseTh2}, we can express the relation between $\mathbb{E} [ || \Tilde{{\bf w}}_{e,n+1} ||_{\text{bvec}^{-1}\{ \boldsymbol{\sigma} \}}^2 ]$ and $\mathbb{E} [ || \Tilde{{\bf w}}_{e,n} ||_{\text{bvec}^{-1}\{ \boldsymbol{\sigma} \}}^2 ]$. We have:
\begin{align}
    \label{SteadyStateMSD}
    \mathbb{E} [ || \Tilde{{\bf w}}_{e,n+1} ||_{\text{bvec}^{-1} \{ \boldsymbol{\sigma} \}}^2 ] &= \mathbb{E} [ || \Tilde{{\bf w}}_{e,n} ||_{\text{bvec}^{-1} \{ \boldsymbol{\sigma} \}}^2 ] \notag \\
    &\hspace{2mm}+ \mathbb{E} [ || \Tilde{{\bf w}}_{e,0} ||_{\text{bvec}^{-1} \{ (\boldsymbol{\mathcal{F}}^{\mathsf{T}} - {\bf I}) (\boldsymbol{\mathcal{F}}^{\mathsf{T}} )^n \boldsymbol{\sigma} \}}^2 ] \notag \\
    &\hspace{2mm} + \mu^2 {\bf h}^{\mathsf{T}} (\boldsymbol{\mathcal{F}}^{\mathsf{T}})^n \boldsymbol{\sigma}.
\end{align}
If we set $\boldsymbol{\sigma} = \text{bvec} \{ \text{blockdiag} \{ {\bf I_D}, {\bf 0}, \hdots, {\bf 0} \} \}$, we obtain the transient expression for the mean square deviation of the global model at iteration $n$: $ \mathbb{E} [ || \Tilde{{\bf w}}_{n} ||^2 ] = \mathbb{E} [ || \Tilde{{\bf w}}_{e,n} ||_{\text{bvec}^{-1} \{ \boldsymbol{\sigma} \}}^2 ]$.

Under the assumptions and conditions for \textbf{Theorem 2}, by letting $n \rightarrow \infty$ in \eqref{baseTh2}, we obtain the expression of the steady-state mean square deviation (MSD) for the PAO-Fed algorithm. We have:
\begin{align}
    \lim\limits_{n \rightarrow \infty} \mathbb{E} [ || \Tilde{{\bf w}}_{e,n} ||_{\text{bvec}^{-1} \{ ({\bf I} -\boldsymbol{\mathcal{F}}^{\mathsf{T}}) \boldsymbol{\sigma} \}}^2 ] = \mu^2 {\bf h}^{\mathsf{T}} \boldsymbol{\sigma}.
\end{align}
By setting $\boldsymbol{\sigma} = ({\bf I} -\boldsymbol{\mathcal{F}}^{\mathsf{T}})^{-1} \text{bvec} \{ \text{blockdiag} \{ {\bf I_D}, {\bf 0}, \hdots, {\bf 0} \} \}$, the  steady-state MSD expression of the global model can be obtained.

\section{Numerical Simulations}

This section demonstrates the performance of the proposed PAO-Fed algorithm through a series of numerical experiments. In these experiments, we compare the performance of the PAO-Fed algorithm with existing methods, specifically, Online-FedSGD, Online-Fed \cite{FedAVG}, and PSO-Fed \cite{Vinay}.

\subsection{Simulation Setup}

We consider a set of $K = 256$ clients connected to a server. Every client has access to imbalanced streaming data. For this purpose, the clients are separated into $4$ data groups for which the progressively available training set are composed of $500$, $1000$, $1500$, and $2000$ samples, respectively. A single data sample is of the form $\{ {\bf x}_{k,n}, y_{k,n} \}$, where this couple is linked by the following nonlinear relation of $\mathbb{R}^4 \longrightarrow \mathbb{R}$:
\begin{align}
    y_{k,n} = &\sqrt{ {\bf x}^2_{k,n}[1] + \sin^2( \pi {\bf x}_{k,n}[4] ) } \\
    &+ (0.8 - 0.5 \exp( - {\bf x}^2_{k,n}[2] ) {\bf x}_{k,n}[3] ) + \eta_{k,n}, \notag
\end{align}
where ${\bf x}_{k,n}[i]$ denotes the $i^{\text{th}}$ element of the vector ${\bf x}_{k,n}$. Note that clients may receive at most one data sample per iteration. The nonlinear model and the data in the space of dimension $L = 4$ are approximated on an RFF-space of dimension $D = 200$.

As discussed in Section III.A, client participation is modeled using the probabilities $p_{k,n}, k \in \mathcal{K}$. Note that a client can only participate at an iteration if it receives new data; otherwise, the probability is set to $0$. The clients of each data group are further separated into $4$ availability groups, dictating their probability $p_{k,n}$ of participating at each iteration. The Bernouilli trial on $p_{k,n}$ dictates if a client is available or not at a given iteration. Unless stated otherwise, the participation probabilities given to the four availability groups are $0.25$, $0.1$, $0.025$, and $0.005$. Finally, each communication to the server will be delayed by more than $l$ iterations with probability $\delta^l, 0<l<l_{\text{max}}$, with, unless stated otherwise, $\delta = 0.2$ and $l_{\text{max}} = 10$. This probability is assumed to be the same for all clients.

The performance of the algorithms is evaluated on a test dataset with the mean squared error (MSE) given at iteration $n$ by:
\begin{align}
    \text{MSE-test} = \frac{1}{\text{MC}} \sum_{e=1}^{\text{MC}} \frac{|| {\bf y}_{\text{test}}^e - ({\bf Z}_{\text{test}}^e)^\mathsf{T} {\bf w}_n^e ||_2^2}{T},
\end{align}
where $\text{MC}$ is the number of Monte Carlo iterations, $T$ is the size of the test dataset, ${\bf y}_{\text{test}}^e$ and ${\bf Z}_{\text{test}}^e$ are the realisation of the data for a given Monte Carlo iteration, and ${\bf w}_n^e$ is the server's model vector for the considered method. When comparing the PAO-Fed algorithm with other methods, the learning rates were set to yield identical initial convergence rates so that steady-state values may be compared. Some algorithms were not able to reach this common convergence rate, but since their steady-state accuracy is lower, comparison is still possible. All the learning rates satisfy the convergence conditions obtained in Section IV for PAO-Fed, and available in \cite{FedAVG,Vinay} for Online-Fed, Online-FedSGD, and PSO-Fed. For instance, in Fig. \ref{PAO_Analysis}, \ref{PAO_Comparison}, and \ref{RealData}, the step-size for the PAO-Fed algorithm is set to $\mu = 0.4$ with $\max\limits_{\forall k, i} \lambda_i({\bf R}_k) = 1.02$.

In the simulations, we implement uncoordinated partial-sharing-based communications from the server to the clients with $\text{diag}({\bf M}_{k,n}) = \text{circshift}(\text{diag}({\bf M}_{1,n}), m k)$ and $\text{diag}({\bf M}_{1,n}) = \text{circshift}(\text{diag}({\bf M}_{1,0}), m n)$. This, in turn, dictates the portion of the model sent by the clients to the server (see Section II C) so that, on average, all portions are equally represented in the aggregation. We recall that $m$ is the number of model parameters shared at each iteration by both the server and the clients, and dictates the communication overhead in partial-sharing-based communications.

We consider different versions of the PAO-Fed algorithm.
\begin{itemize}
    \item \textbf{PAO-Fed-C2} and \textbf{PAO-Fed-U2} are the proposed PAO-Fed method using coordinated and uncoordinated partial-sharing, respectively. Their selection matrices evolve as described in Section II C. That is, they use the local \eqref{ClientUpdateSelected} and autonomous local \eqref{LocalUpdateRe} updates to share a better-refined model. Further, as seen in \eqref{ServerUpdate}, they possess a weight-decreasing mechanism with $\alpha_l = 0.2^l, 0 \leqslant l \leqslant l_{\text{max}}$.
    \item \textbf{PAO-Fed-C1} and \textbf{PAO-Fed-U1} use coordinated and uncoordinated partial sharing, with selection matrices evolving as described in Section II C. However, they do not possess a weight-decreasing mechanism, that is, $\alpha_l = 1, 0 \leqslant l \leqslant l_{\text{max}}$.
    \item \textbf{PAO-Fed-C0} and \textbf{PAO-Fed-U0} use coordinated and uncoordinated partial-sharing, respectively, and do not possess a weight-decreasing mechanism. They differ from the previous versions in that clients share the last received server model, refined once by the local update process (e.g., in Fig. \ref{PS} ${\bf w}_{k,n+1}(0)$ would be shared instead of ${\bf w}_{k,n+1}(1)$). They can be seen as performing OnlineFedSGD on a rolling portion of the model, without further tuning.
\end{itemize}
Unless specified otherwise, every PAO-Fed-based algorithm is set to share $m = 4$ model parameter per communication round, leading to a reduction in communication by $98 \%$. 


\subsection{Hyper Parameters Selection}

\begin{figure*}[t!]
    \centering
    \subfigure[]{\includegraphics[width=0.31\textwidth]{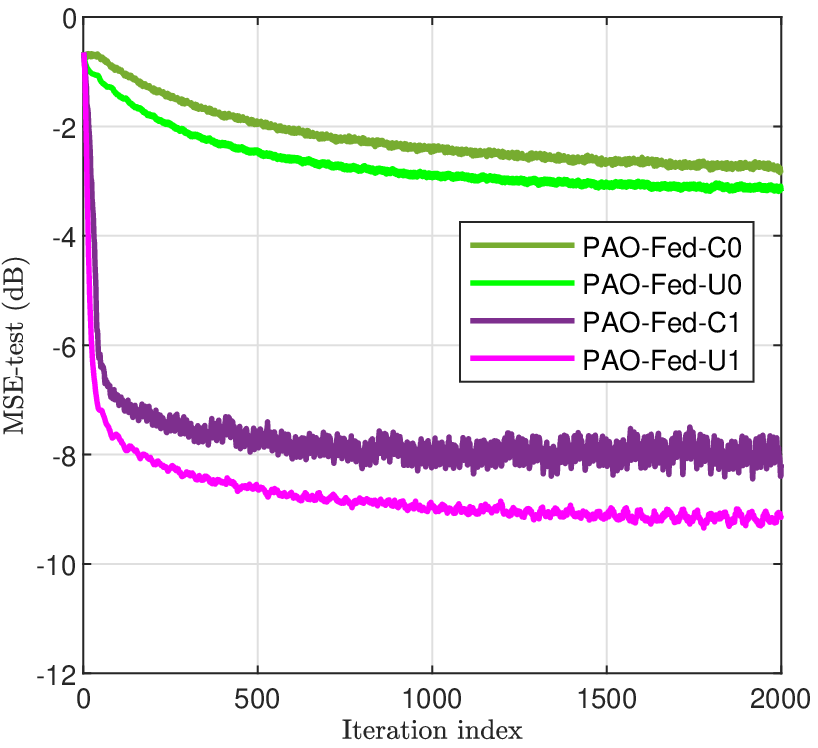}}
    \subfigure[]{\includegraphics[width=0.31\textwidth]{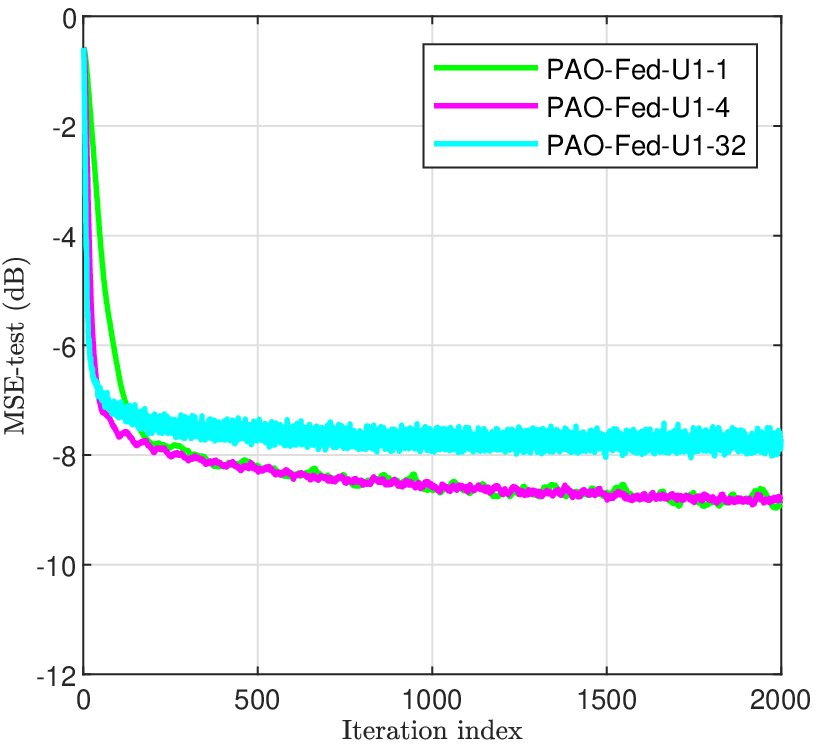}}
    \subfigure[]{\includegraphics[width=0.31\textwidth]{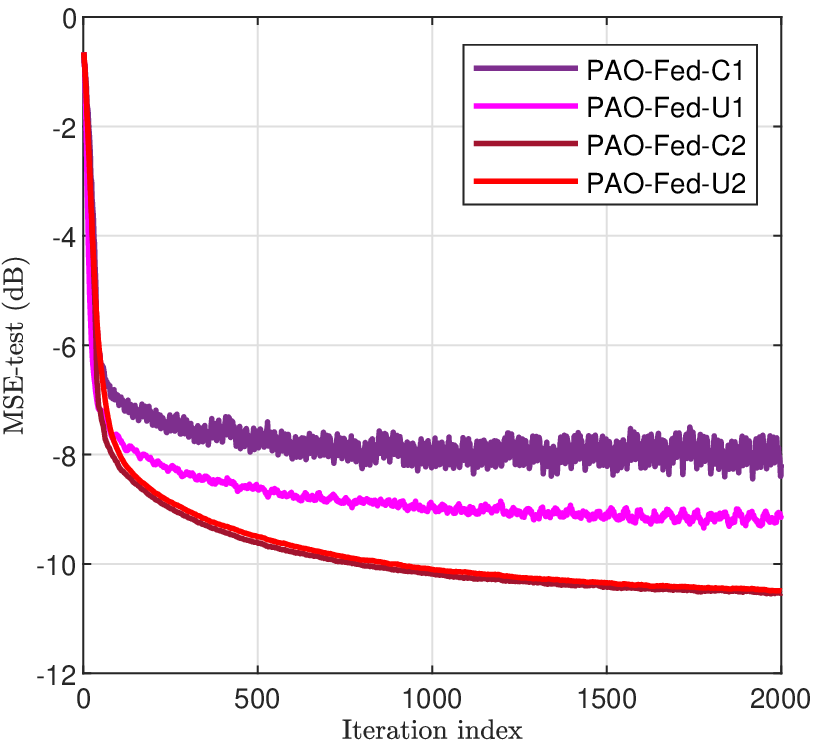}}
    \caption{Optimization of the PAO-Fed method. (a) the use of local updates and coordinated/uncoordinated partial-sharing, (b) communication overhead, (c) weight-decreasing mechanism for delayed updates.}
    \label{PAO_Analysis}
\end{figure*}

In the first experiments, we study the impact of the hyper parameters on the convergence properties of the PAO-Fed algorithm. Specifically, we investigate the impact of the choice of the selection matrices, the number of model parameters shared, and the scale of the weight-decreasing mechanism for delayed updates. The corresponding learning curves in Fig. \ref{PAO_Analysis} display the $\text{MSE-test}$ in dB versus the iteration index.

Firstly, we examined how the choice of the selection matrices ${\bf M}_{k,n}$ and ${\bf S}_{k,n}$ impact the convergence properties of the PAO-Fed algorithm. These matrices select the model portion to be shared between the server and clients (see Section II C). The versions PAO-Fed-(C/U)0 are set with ${\bf S}_{k,n} = {\bf M}_{k,n}$; that is, the last received portion from the server is updated once by the local learning process at the clients before being sent back to the server. On the contrary, the versions PAO-Fed-(C/U)1 are set as in \eqref{SelectionMatrixChoice} and \eqref{SelectionMatrixChoice2}; that is, the received portions from the server will be updated several times by the local learning process to accumulate information, in a manner similar to batch learning, before being sent back to the server. We observe in Fig. \ref{PAO_Analysis} (a) that the versions PAO-Fed-(C/U)1 outperform the versions PAO-Fed-(C/U)0. For this reason, we will only consider the versions of the PAO-Fed algorithm making full use of the local updates in the following. We also notice in this experiment that it is best to use uncoordinated partial-sharing in asynchronous settings, this contradicts the behavior of partial-sharing-based communications in ideal settings, where coordinated partial-sharing performs slightly better than uncoordinated, as explained in \cite{Vinay}.


Secondly, we studied the impact of the number $m$ of model parameters shared by participating clients and the server during the learning process. Fig. \ref{PAO_Analysis} (b) shows the performance of the PAO-Fed-U1 algorithm (uncoordinated, making use of local updates) for different values of $m$, namely $m=1$, $m=4$, and $m=32$. Although sharing more model parameters increases the initial convergence speed, we observed that it decreases the final accuracy for larger $m$ values. This contradicts previous results in the literature about the behavior of partial-sharing in ideal settings \cite{Vinay}. In fact, sharing more model parameters increases the potential negative impact of one single delayed update carrying outdated information, decreasing the overall accuracy. Sharing a small number of model parameters limits the impact of a given update, providing some level of protection against outdated information, and ensuring better model fitting. We chose to set $m=4$ as a baseline, as it presents a good compromise between initial convergence speed, steady-state accuracy, and communication overhead reduction.


Finally, we tried to attenuate the harmful effect of delayed updates on the convergence properties of the algorithm. To this aim, the weight-decreasing mechanism for delayed updates proposed in \eqref{ServerUpdate} is set for the versions PAO-Fed-(C/U)2 with $\alpha_l = 0.2^l, 0 \leqslant l \leqslant l_{\text{max}}$. In Fig. \ref{PAO_Analysis} (c), we display the performance of these methods alongside PAO-Fed-(C/U)1. We observe that decreasing the weight of the delayed updates significantly improves the performance of the PAO-Fed algorithm on the considered asynchronous settings. The proposed mechanism considers the relevance of delayed and potentially outdated updates by effectively reducing their impact on the server model, especially for substantial delays. By doing so, the negative effect of delayed updates is mitigated; in particular, when using the aforementioned weight-decreasing mechanism, the performance difference between coordinated and uncoordinated partial-sharing vanishes (see PAO-Fed-C2 versus PAO-Fed-U2 in the figure). 

\subsection{Comparison of PAO-Fed with Existing Algorithms}

\begin{figure*}[t!]
    \centering
    \subfigure[]{\includegraphics[width=0.31\textwidth]{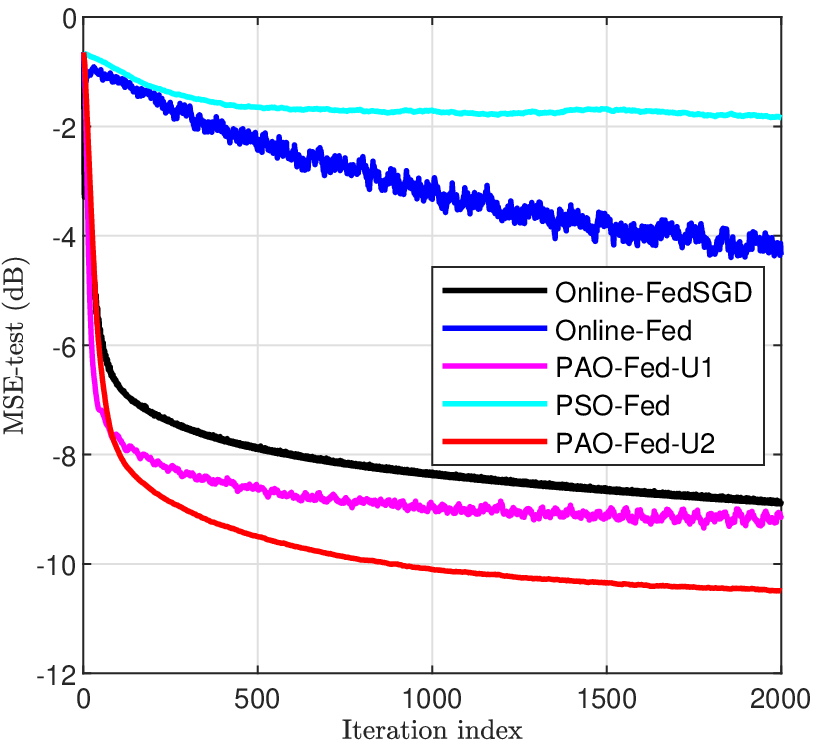}}
    \subfigure[]{\includegraphics[width=0.31\textwidth]{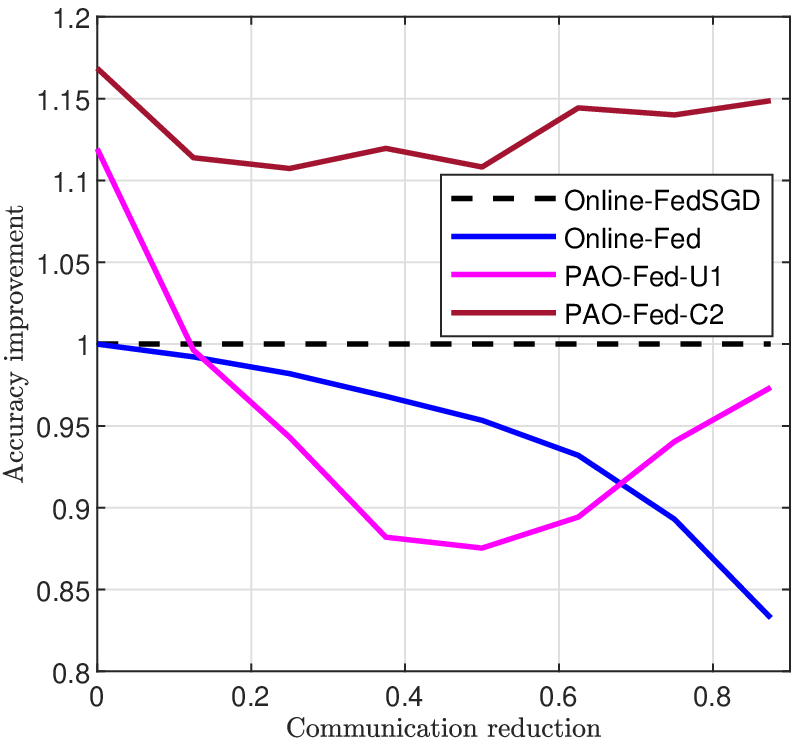}}
    \subfigure[]{\includegraphics[width=0.30\textwidth]{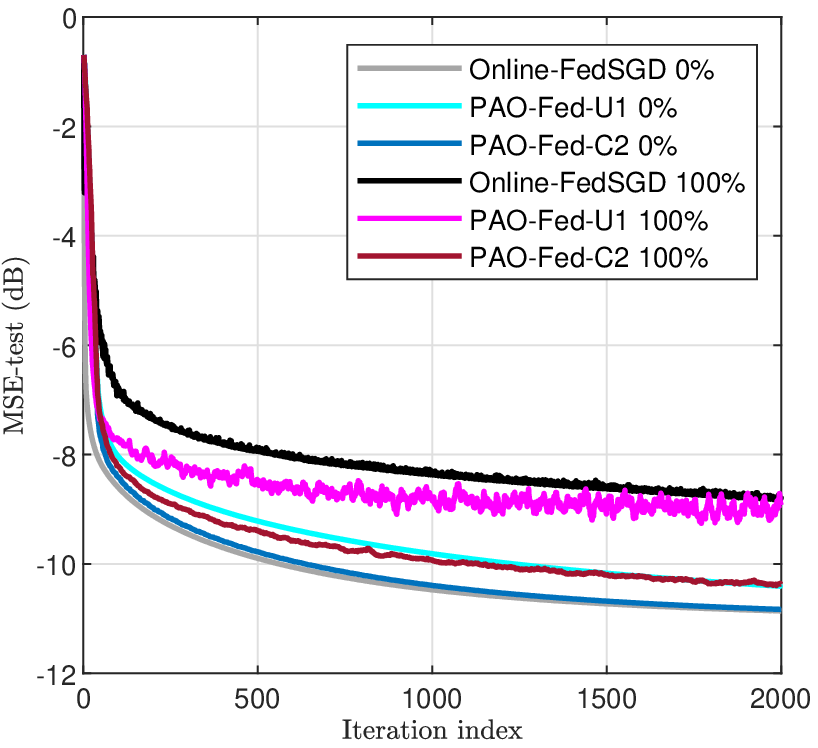}}
    \caption{Comparison of PAO-Fed with existing methods. (a) learning curves, (b) communication reduction vs accuracy, (c) impact of straggler clients.}
    \label{PAO_Comparison}
\end{figure*}

In the following experiments, we compare the performance of the PAO-Fed algorithm with existing online FL methods in the literature. Fig. \ref{PAO_Comparison} (a) and (c) display the $\text{MSE-test}$ in dB versus the iteration index, and Fig. \ref{PAO_Comparison} (b) displays accuracy variation versus communication overhead reduction.


Firstly, we compared PAO-Fed-U1 and PAO-Fed-U2 with PSO-Fed \cite{Vinay}, Online-Fed \cite{FedAVG}, and Online-FedSGD. The corresponding learning curves are displayed in Fig. \ref{PAO_Comparison} (a). First, we observe that Online-Fed and PSO-Fed both perform poorly; sub-sampling the already reduced pool of available clients is not a viable solution to reduce communication overhead in asynchronous settings. Then, we observe that both PAO-Fed-U1 and PAO-Fed-U2 outperform Online-FedSGD while using $98\%$ less communication overhead. The reason for this very good performance is twofold. First, using the local and autonomous local updates in the PAO-Fed algorithm allows it to extract more information from the sparsely participating clients. Second, partial-sharing-based communication provides the PAO-Fed algorithm with an innate resilience to the negative impact of delayed updates; this resilience is further increased in the PAO-Fed-U2 algorithm with the weight-decreasing mechanism, hence its better performance.


Secondly, to illustrate the benefits of the proposed method, we display the communication reduction versus the accuracy improvement in Fig. \ref{PAO_Comparison} (b). To this aim, we use Online-FedSGD as a baseline for accuracy since it communicates all parameters in each iteration. On the abscissa, we display the communication reduction ratio; $0$ corresponds to no communication reduction, and $0.8$ corresponds to an $80\%$ decrease in the communication overhead. On the ordinate, we display the accuracy improvement over Online-FedSGD after 2000 iterations, e.g., $1.1$ corresponds to a $10\%$ reduction of the error. First, we notice that the PAO-Fed methods start with higher accuracy without communication overhead reduction. This is due to the use of the autonomous local updates \eqref{LocalUpdateRe} by those methods, and the weight-decreasing mechanism of PAO-Fed-C2. Second, we observe that reducing the communication overhead with scheduling comes with an exponential accuracy cost. Third, although partial-sharing-based communications initially comes at a high accuracy cost, this cost is reversed as the message length diminishes because of the innate resilience against delayed updates and the use of the local and autonomous local updates. Last, we notice that the weight-decreasing mechanism of the PAO-Fed-C2 algorithm allows it to outperform every other method no matter the reduction in communication overhead.


Finally, in Fig. \ref{PAO_Comparison} (c), we observe the impact of the straggler clients on the convergence properties of the algorithms. To this aim, we compare the performance of the algorithms in the proposed settings ($100\%$ of clients are potential stragglers) to their performance in an ideal setting where clients are always available when they receive new data and their communication channels do not suffer from delays ($0\%$ of clients are potential stragglers). We observe that, in the absence of straggler clients, the methods using coordinated partial-sharing achieve greater accuracy, almost identical to methods with no communication overhead reduction, while the methods using uncoordinated partial-sharing have slightly worse performance, this corresponds to the results obtained in \cite{Vinay}. Furthermore, we see that the PAO-Fed-C2 algorithm used on straggler clients has convergence properties almost similar to the ones of algorithms in a perfect setting.

\subsection{Performance on a Real-world Dataset}

\begin{figure}
    \centering
    \includegraphics[width=0.35\textwidth]{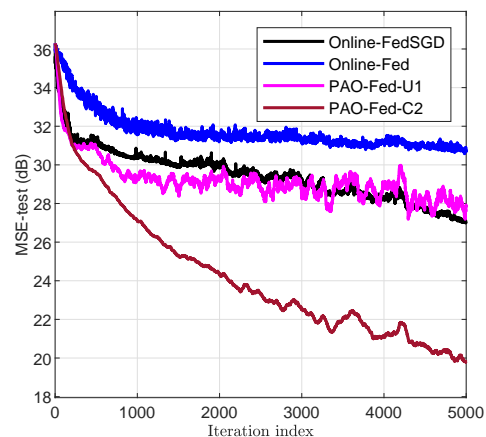}
    \caption{Learning curves on the \textit{bottle} dataset.}
    \label{RealData}
\end{figure}

\begin{figure*}
    \centering
    \subfigure[]{\includegraphics[width=0.30\textwidth]{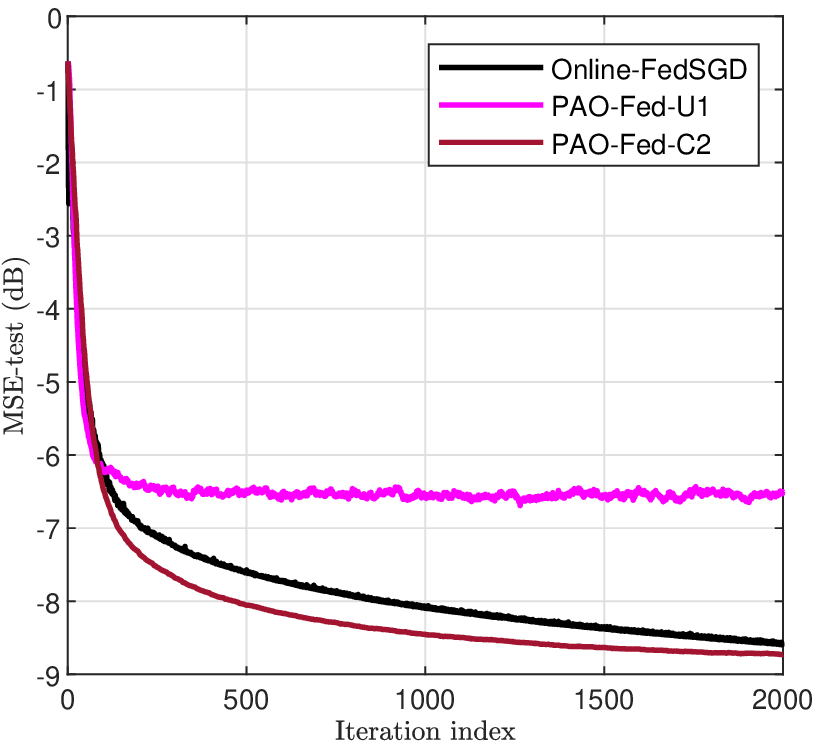}}
    \subfigure[]{\includegraphics[width=0.30\textwidth]{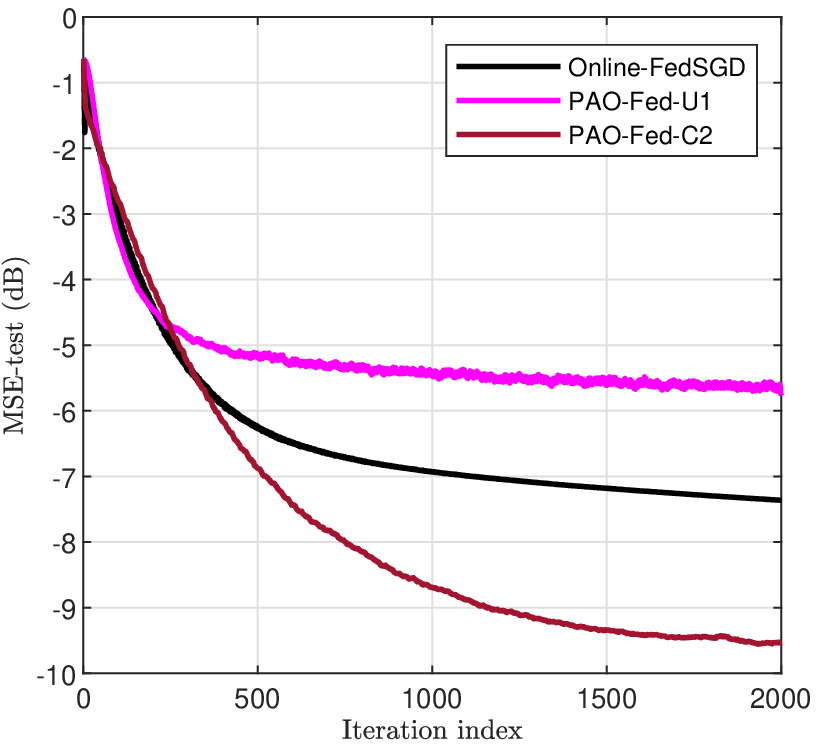}}
    \subfigure[]{\includegraphics[width=0.31\textwidth]{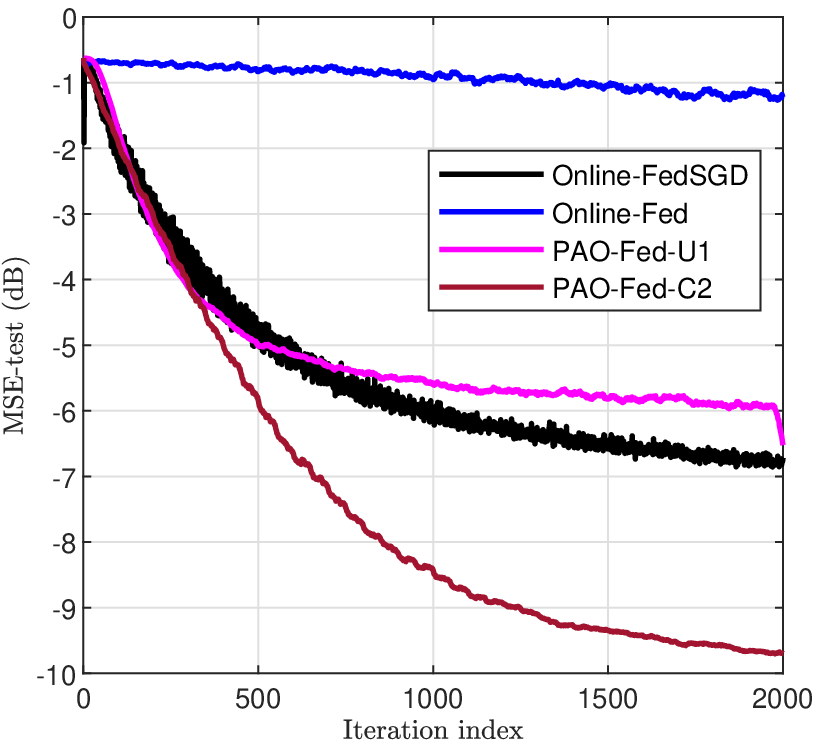}}
    \caption{Learning curves in different environments. (a) full server communication, (b) common delays, (c) advanced straggler behavior.}
    \label{Altered_Settings}
\end{figure*}

In order to demonstrate the usability of the proposed solution further, we conducted simulations on a real-world dataset. The corresponding learning curves are plotted in Fig. \ref{RealData}. We used the \textit{bottle} dataset from CalCOFI \cite{CalCOFI}, composed of more than 800,000 samples of water with various parameters such as temperature, salinity, O2 saturation, etc. The salinity of the water is linked in a nonlinear manner to the other available parameters; we will use the proposed method to learn the salinity level from the other parameters. For the purpose of the experiment, we consider only 80,000 samples that we distribute progressively and unevenly to the clients throughout the learning process. Further, we simulate the straggler-like behavior of the clients as mentioned above (availability groups are $0.25$, $0.1$, $0.025$, and $0.005$; each communication to the server will be delayed by more than $l$ iterations with probability $\delta^l, 0<l<l_{\text{max}}$, with $\delta = 0.2$ and $l_{\text{max}} = 10$). We observe identical results to the simulations on simulated data - the PAO-Fed-U1 algorithm is able to achieve the same accuracy as Online-FedSGD while using $98\%$ less communications, and the PAO-Fed-C2 algorithm, also using $98\%$ less communications, is able to outperform all other methods.

\subsection{Impact of the Environment on Convergence Properties}


In these last experiments, we study the impact that a change in the external environment can have on the convergence properties of the proposed PAO-Fed algorithm and existing methods. The corresponding learning curves in Fig. \ref{Altered_Settings} display the $\text{MSE-test}$ in dB versus the iteration index.


Firstly, we studied in Fig. \ref{Altered_Settings} (a) the importance of using partial-sharing-based communications both at the server and at the clients. The algorithms using partial-sharing-based communications have been altered in this simulation with ${\bf M}_{k,n} = {\bf I}, \forall k,n$; that is, the server sends its entire model to the participating clients at each iteration. This modification can be appealing if the server is not subject to power constraints. The clients behave normally and only send a portion of their local model; however, unlike in the other simulations, the received global model replaces the local model at each participant, see \eqref{ClientUpdateSelected}. In such a case, we observe that the performance of the partial-sharing-based methods is drastically reduced. It is the information kept by the clients in the not-yet-shared portions of their local models that allow partial-sharing-based methods to outperform Online-FedSGD. We note that clients may choose to ignore part of the received model to avoid this downfall.


Secondly, we studied the algorithm behaviors in an environment where most communications are delayed, but delays cannot be too lengthy. To this aim, the delay probability has been significantly increased and the maximum possible delay reduced ($\delta = 0.8$ and $l_{\text{max}} = 5$). We observe in Fig. \ref{Altered_Settings} (b) that the limited maximum delay allows Online-FedSGD to outperform PAO-Fed-U1, as the benefit of partial-sharing against data of poor quality does not out-weight the smaller amount of communication available to PAO-Fed-U1. To compensate for the fact that most incoming information is weighted down by the weight-decreasing mechanism of PAO-Fed-C2, its learning rate has been increased to near its maximum value obtained in \textbf{Theorem 2}. Despite this, the PAO-Fed-C2 algorithm reaches very low steady-state error and significantly outperforms Online-FedSGD. 


Finally, we modeled an environment where availability groups are given the probabilities $0.025$, $0.01$, $0.0025$, and $0.0005$; communications to the server have a probability $\delta = 0.4$ to be delayed. Further, delays last for more than $l$ iterations, $l$ taking the values $10 i, 0 \leqslant i \leqslant 6$, with probability $\delta^{\frac{l}{10}}$; $l_{\text{max}}$ is set to $60$. This notably implies that, in this environment, delayed updates have a greater probability of arriving after a non-delayed update coming from the same client. Such an environment where clients are less likely to be available to participate, communications are more likely to be delayed, and delays last for more iterations, is less favorable to learning. An application relying on edge devices that are poorly available and unreliable would evolve in an environment similar to this. Fig. \ref{Altered_Settings} (c) presents the learning curves of Online-Fed, Online-FedSGD, and the PAO-Fed algorithm in this new environment to see how it may impact the convergence properties of the algorithms. We observe that, in this environment, reducing the weight given to the delayed updates gains importance as the accuracy difference between PAO-Fed-C2 and PAO-Fed-U1 increases. In fact, delayed updates may carry information that is significantly outdated and, therefore, prevent the algorithms not using a weight-decreasing mechanism for delayed updates to reach satisfactory steady-state error. For this reason, the PAO-Fed-C2 algorithm achieves significantly better accuracy than Online-FedSGD in this environment.

\section{Conclusions}
This paper proposed an energy-efficient federated learning (FL) algorithm adapted to a realistic environment. The proposed FL algorithm operates with significantly reduced communication requirements and can cope with an unevenly distributed system with poor client availability, potential failures, and communication delays. The proposed partial sharing mechanism reduces the communication overhead and diminishes the negative impact of delayed updates on accuracy. We further propose a weight-decreasing aggregation mechanism that emphasizes more recent updates to improve performance in environments suffering from substantial delays, poor participation, and straggler devices. Our numerical results show that the proposed algorithm outperforms standard FL methods in such an environment while reducing the communication overhead by $98$ percent.

\begin{appendices}

\section{Evaluation of $\mathbb{E} [ {\bf A}_{e,n}]$ and $\mathbb{E} [ {\bf B}_{e,n}]$}
The matrix ${\bf A}_{e,n}$ is composed of $D \times D$-sized blocks ${\bf A}_{i,j,n}$, given by:
\begin{align}
    {\bf A}_{i,j,n} =
    \begin{cases}
        {\bf I}_D &\text{if} \; i=j \land (i=1 \lor i > K+1), \notag \\
        a_{k,n} {\bf M}_{k,n} &\text{if} \; i \in [|2, \hdots, K+1|] \land j = 1, \notag \\
        {\bf I}_D - a_{k,n} {\bf M}_{k,n} &\text{if} \; i \in [2, \hdots, K+1] \land i=j, \notag \\
        {\bf 0}_D &\text{otherwise}, \notag
    \end{cases}
\end{align}
where $k = i-1$.

We note that $\mathbb{E} [a_{k,n} {\bf M}_{k,n}] = p_{k,n} p_m {\bf I}_D$, with $p_{k,n}$ being the probability that client $k$ participates at iteration $n$, and $p_m$ being the probability that a given model parameter is selected by the selection matrix (i.e., the density of the selection: $\frac{m}{D}$). Since $0 \leqslant p_{k,n} p_m \leqslant 1$, and given the above decomposition, matrix $\mathbb{E} [ {\bf A}_{e,n}]$ is right stochastic.

Further, we note that by construction, $(a_{k,n} {\bf M}_{k,n})^2 = a_{k,n} {\bf M}_{k,n}$; therefore, under \textbf{Assumption 3}, we have:
\begin{align}
    \mathbb{E} &[a_{k,n} {\bf M}_{k,n} a_{k',n'} {\bf M}_{k',n'}]  \notag \\
    &=\begin{cases}
        p_{k,n} p_m {\bf I}_D &\text{if} \; k = k' \land n = n', \notag \\
        p_{k,n} p_{k',n'} p_m ^2 {\bf I}_D &\text{otherwise}. \notag
    \end{cases}
\end{align} \\

Similarly, we decompose the matrix ${\bf B}_{e,n}$ in $D \times D$-sized blocks ${\bf B}_{i,j,n}$ as follows:
\begin{align}
    {\bf B}_{i,j,n} 
    =\begin{cases}
        {\bf B}_{n} &\text{if} \; i=j=1, \notag \\
        {\bf B}_{0,n}^{(j-1)} &\text{if} \; i=1 \land j \in [2, K+1], \notag \\
        {\bf B}_{\ceil*{\frac{j-1}{K}}-3,n}^{(j - 1 \mod K)} &\text{if} \; i=1 \land j \in \notag \\
        &[|3K+2, \hdots, (l_{\text{max}}+3)K+1|], \notag \\
        {\bf I}_D &\text{if} \; i \in [|1,2|] \land j=2, \notag \\
        {\bf I}_D &\text{if} \; i>3 \land j>2 \land i=j+1, \notag \\
        {\bf 0}_D &\text{otherwise}. \notag
    \end{cases},
\end{align}
The blocks are given by:
\begin{align}
     {\bf B}_{n} &= {\bf I} - \sum_{l=0}^{l_{\text{max}}} \alpha_l \sum_{k \in \mathcal{K}_{n,l}} \frac{b_{k,n,l} }{|\mathcal{K}_{n,l}|} {\bf S}_{k,n-l}, \notag \\
     {\bf B}_{l,n}^{(k)} &= {\bf B}_{l,n}[k], \notag \\
     {\bf B}_{l,n} &= \left[ \frac{\alpha_l b_{1,n,l}}{|\mathcal{K}_{n,l}|} {\bf S}_{1,n-l}, \dotsb, \frac{\alpha_l b_{K,n,1}}{|\mathcal{K}_{n,l}|} {\bf S}_{K,n-l} \right]. \notag
\end{align}

We note that by construction,
\begin{align}
    {\bf B}_{n} + \sum_{l=1}^{l_{\text{max}}} \sum_{k=1}^{K} {\bf B}_{l,n}^{(k)} = {\bf I}, \notag
\end{align}
hence, the matrix $\mathbb{E} [ {\bf B}_{e,n} ]$ is right stochastic as well.

\newpage
\section{Evaluation of $\boldsymbol{\mathcal{Q}}_{{\bf A}}$ and $\boldsymbol{\mathcal{Q}}_{{\bf B}}$}

We decompose matrix $ \boldsymbol{\mathcal{Q}}_{{\bf A}} $ into $D \times D$-sized blocks and prove the property by computing the Kronecker product $ {\bf A}_{e,n} \otimes_b {\bf A}_{e,n} $ before taking the expectation. In particular, we have:
\begin{align}
    \boldsymbol{\mathcal{Q}}_{{\bf A}} = [ \mathbb{E} [{\bf A}_{i,j,n} \otimes_b {\bf A}_{e,n}], (i,j) \in [|1,\hdots,K (l_{\text{max}} + 1) + 1|]^2], \notag
\end{align}
and we note that $ \boldsymbol{\mathcal{Q}}_{{\bf A}} $ can be proven to be right stochastic one block-row at a time, considering sets of $D$ rows indexed by $i$ in the above equation.

The property is easy to prove on the block-rows $i = 1$ and $ i > K+1$. On those block-rows, we have:
\begin{align}
    {\bf A}_{i,j,n} =
    \begin{cases}
        {\bf I}_D &\text{if} \; i=j \notag \\
        {\bf 0}_D &\text{otherwise} \notag
    \end{cases},
\end{align}
therefore, since $\mathbb{E} [{\bf A}_{e,n}]$ satisfy the property, it is satisfied on those block-rows. 

We now consider the remaining block-rows. For this purpose, let $i \in [|2, \hdots, K+1|]$. According to the decomposition of the left-hand side ${\bf A}_{e,n}$, the block-row $i$ of $\boldsymbol{\mathcal{Q}}_{{\bf A}}$ reduces to only two non-zero elements, $\mathbb{E} [{\bf A}_{i,1,n} \otimes_b {\bf A}_{e,n}]$ and $\mathbb{E} [{\bf A}_{i,i,n} \otimes_b {\bf A}_{e,n}]$. Hence we can compute:
\begin{align}
    \mathbb{E} &[{\bf A}_{i,1,n} \otimes_b {\bf A}_{e,n}] + \mathbb{E} [{\bf A}_{i,i,n} \otimes_b {\bf A}_{e,n}] \notag \\
    &= \mathbb{E} [a_{i-1,n} {\bf M}_{i-1,n} \otimes_b {\bf A}_{e,n}] \notag \\
    &+ \mathbb{E} [({\bf I}_D - a_{i-1,n} {\bf M}_{i-1,n}) \otimes_b {\bf A}_{e,n}] \notag \\
    &= \mathbb{E} [{\bf I}_D \otimes_b {\bf A}_{e,n}], \notag
\end{align}
and conclude that the block-row $i$ satisfies the property. \\

Similarly, we decompose the matrix $\boldsymbol{\mathcal{Q}}_{{\bf B}}$ into $D \times D$-sized blocks and prove that it is right stochastic by computing the Kronecker product ${\bf B}_{e,n} \otimes_b {\bf B}_{e,n}$ before taking the expectation. We have:
\begin{align}
    \boldsymbol{\mathcal{Q}}_{{\bf B}} = [ \mathbb{E} [{\bf B}_{i,j,n} \otimes_b {\bf B}_{e,n}], (i,j) \in [1,\hdots,K (l_{\text{max}} + 1) + 1]^2]. \notag
\end{align}

The evaluation is trivial for the block-rows $i \in [2,\hdots,K (l_{\text{max}} + 1) + 1]$, where the decomposition of the left-hand side ${\bf B}_{e,n}$ reduces to only one non-zero element: ${\bf I}$. Therefore, since ${\bf B}_{e,n}$ satisfies the property, it is satisfied on those block-rows.

We now consider the block-row $i = 1$ and compute the sum of the elements as:
\begin{align}
     &\mathbb{E} [ {\bf B}_{n} \otimes_b {\bf B}_{e,n} +  \sum_{l=1}^{l_{\text{max}}} \sum_{k=1}^{K} {\bf B}_{l,n}^{(k)} \otimes_b {\bf B}_{e,n} ] \notag \\
     &= \mathbb{E} [ {\bf I}_D \otimes_b {\bf B}_{e,n} ], \notag
\end{align}
by construction of the ${\bf B}_{l,n}^{(k)}$ matrices. We conclude that the block-row $i = 1$ satisfies the property as well.

We have proven that both $\boldsymbol{\mathcal{Q}}_{{\bf A}}$ and $\boldsymbol{\mathcal{Q}}_{{\bf B}}$ are right stochastic matrices.

\end{appendices}

\bibliographystyle{IEEEtran}
\bibliography{Federated}

\end{document}